# Taking it further: leveraging pseudo labels for field delineation across label-scarce smallholder regions

Philippe Rufin[1,2*], Sherrie Wang[3], Sá Nogueira Lisboa[4,5], Jan Hemmerling[6],

Mirela G. Tulbure[7], Patrick Meyfroidt[1,8]

[1] Earth and Life Institute, UCLouvain, Place Pasteur 3, 1348 Louvain-la-Neuve, Belgium

[2] Geography Department, Humboldt-Universität zu Berlin, Unter den Linden 6, 10117 Berlin, Germany

[3] Department of Mechanical Engineering, Massachusetts Institute of Technology, 77 Massachusetts Avenue, Cambridge MA 02139-4307, USA

[4] Nitidae, 500, rue Jean-François Breton, 34 000 Montpellier - France

[5] Department of Forestry Engineering, Universidade Eduardo Mondlane, Av. Julius Nyerere, Maputo, Mozambique

[6] Thünen-Institute of Farm Economics, Bundesallee 63, 38116 Braunschweig, Germany

[7] Center for Geospatial Analytics, North Carolina State University, Jordan Hall, 5112, 2800 Faucette Dr, Raleigh, NC 27695, USA

[8] F.R.S.-FNRS, Rue d'Egmont 5, 1000 Brussels, Belgium

* corresponding author

# Abstract

Satellite-based field delineation has entered a quasi-operational stage due to recent advances in machine learning for computer vision. Transfer learning allows for the resource-efficient transfer of pre-trained field delineation models across heterogeneous geographies. However, the scarcity of labeled data for complex and dynamic smallholder landscapes, particularly in Sub-Saharan Africa, remains a major bottleneck. This study explores opportunities for using pre-trained models to generate sparse (i.e. not fully annotated) field delineation pseudo labels for fine-tuning models across geographies and sensor characteristics. We build on a FracTAL ResUNet trained for crop field delineation in India (median field size of 0.24 ha) based on multi-spectral imagery at 1.5 m spatial resolution. We use this pre-trained model to generate pseudo labels for model training in smallholder landscapes of Mozambique (median field size of 0.06 ha) based on sub-meter resolution true-color satellite imagery. We designed multiple pseudo label selection strategies based on field-level probability scores and compared the quantities, area properties, seasonal distribution, and spatial agreement of the pseudo labels against human-annotated training labels (n = 1,512). We then used the human-annotated labels and the pseudo labels for model fine-tuning and compared predictions against human field annotations (n = 2,199). We evaluated performance with regards to object-level spatial agreement and site-level field size estimation. Our results indicate i) a good baseline performance of the pre-trained model in both field delineation (mean intersection over union (mIoU) of 0.634) and field size estimation (mean root mean squared error (mRMSE) of 0.071 ha), and ii) the added value of regional fine-tuning with performance improvements in nearly all experiments (mIoU increases of up to 0.060, mRMSE decreases of up to 0.034 ha). Moreover, we found iii) substantial performance increases when using only pseudo labels (up to 77% of the mIoU increases and 68% of the mRMSE decreases obtained by human labels), and iv) additional performance increases (mIoU 0.008, mRMSE: -0.003 ha) when complementing human annotations with pseudo labels. Pseudo labels can be efficiently generated at scale and thus facilitate domain adaptation in label-scarce settings. The workflow presented here is a stepping stone for overcoming the persisting data gaps in heterogeneous smallholder agriculture of Sub-Saharan Africa, where labels are commonly scarce.

# Introduction

Smallholder farms operating on less than 2 ha of land produce an estimated 30–34% of the global food supply on 24% of the gross agricultural area (Ricciardi et al., 2018), have higher land productivity (Chiarella et al., 2023), and harbor greater biodiversity as compared to larger farms (Ricciardi et al., 2021). On the other hand, smallholder farming still constitutes a large share of agriculture-driven deforestation across the tropics (FAO, 2023). The commonly used dichotomy between small and large farms, however, tends to overlook the ongoing structural dynamics related to medium-scale farms (5-100 ha) (Meyfroidt, 2017), which represent an increasingly large portion of agricultural operations in Sub-Saharan Africa (Jayne et al., 2022), causing dynamics of farm consolidation with yet weakly understood impacts and spill-overs affecting productivity, land availability, and labor opportunities at the regional level. As such, a more detailed understanding of land management and change trajectories in smallholder-dominated contexts is required to better balance trade-offs between the social, environmental, and economic impacts of agriculture (Chiarella et al., 2023).

Quantitative and spatially detailed information is limited for tropical smallholder agriculture, and Sub-Saharan Africa in particular (Whitcraft et al., 2019). Numerous empirical blind spots persist, including spatially explicit data on dominant crop types or farming systems (Lambert et al., 2018) or plot-level estimates of productivity (Burke and Lobell, 2017). Moreover, the spatial distribution of agricultural fields and their sizes remain unknown at the sub-national level, hampering a more detailed understanding of the ongoing structural transformations of agriculture in Sub-Saharan Africa (Burke et al., 2020; Jayne et al., 2022; Liverpool-Tasie et al., 2020).

The combination of Earth Observation (EO) data and deep learning offers great potential to close some of the knowledge gaps pertaining to field size distribution by producing accurate, timely, and spatially detailed information in complex smallholder environments (Nakalembe and Kerner, 2023). Notable progress has been made in delineating smallholder fields in EO data (Mei et al., 2022; Persello et al., 2019). Challenges related to the spatially fragmented and dynamic nature of smallholder agriculture with region-specific crop portfolios, landscape structures, and sub-plot land management have been successfully addressed at the local level by using selected very-high-resolution (VHR) EO data (Jong et al., 2022; Long et al., 2022; Persello et al., 2019). The development of regional to continental scale field delineations in Africa, however, is – amongst other factors - severely challenged by a reference data bottleneck (Nakalembe and Kerner, 2023). Label acquisition has to be conducted individually for each target region since cadaster or land parcel information is typically not available in smallholder regions. Further, state-of-the-art methods in deep learning require semantically complete field instances as reference data (Cai et al., 2023; Li et al., 2023; Long et al., 2022; Waldner and Diakogiannis, 2020), which can only be collected by well-trained personnel and with repeated expert quality screening, which makes it cost-intensive and often limits the quantities of available labels (Estes et al., 2022).

In recent years, a wealth of training strategies has been proposed to address the paucity of reference data inherent to EO applications (Safonova et al., 2023). Transfer learning, i.e. leveraging pre-trained models, provides a promising avenue for field delineation in data-constrained smallholder settings. Previous studies suggested training models in (source) regions with ubiquitous labels and fine-tuning the models for use in another (target) region using limited reference data (Kerner et al., 2023; Wang et al., 2022). These studies revealed i) equal or even increased performance when applying fine-tuned models for use in a new target region compared to training models from scratch, and ii) significant reductions in training data requirements. For example, Wang et al. (2022) obtained the highest performance with only 500 - 1,000 labeled fields by combining weak supervision (i.e., partially labeled image chips) with transfer learning. Kerner et al. (2023) used a model fine-tuned for field delineation in South Africa and demonstrated that satisfactory performance (mean intersection over union (mIoU) = 0.64) could be achieved for smallholder field delineation in Kenya without regional fine-tuning.

The potential of transfer learning for task-specific geographic transfer is thus large. The crucial aspect in this regard is domain adaptation between the source and target domain, such as differences in geographic realms (i.e. field sizes, shapes, texture, appearance of field boundaries, or seasonal variations) and image characteristics (i.e. spatial resolution, viewing angle, spectral bands, radiometric properties). A variety of approaches address domain adaptation aiming either at reducing label requirements (e.g. transfer learning, active learning, or semi-supervised learning) or entirely automating domain adaptation without the need for any training labels in the target domain (e.g. self-supervised learning, or self-training; see Toldo et al. (2020) for a review of approaches suitable for dense (i.e. pixel-level) predictions, commonly referred to as semantic segmentation). In the case of semantic segmentation, previous studies approached domain adaptation by optimizing for additional learning objectives during training ("online"), such as image-level test time entropy minimization (Wang et al., 2020), or the use of self-training by complementing source data with pseudo labels, which represent predictions on unlabeled data in the target domain (Lee, 2013; Prabhu et al., 2022; Zou et al., 2021).

Self-training based on pseudo labels poses a particularly useful avenue for transfer learning with weak supervision, which may be mobilized without formulating additional learnable objectives in model architectures (i.e. "offline"), and in a source-free setting (i.e. where the training data of the pre-trained model is not readily available). Large quantities of sparse (i.e. partially annotated) labels can be efficiently generated using pre-trained models given unlabeled EO data, which may support domain adaptation. Importantly, however, self-training involves a risk for error propagation with consequent performance loss, e.g. when erroneous pseudo labels are considered during training. Thus, approaches for selecting accurate and informative labels are essential, and careful performance evaluation and testing for domain-specific biases should be conducted (Mei et al., 2020; Prabhu et al., 2022). The selection of pseudo labels using constant confidence thresholds typically favors easy cases or classes and generate non-diverse and potentially biased pseudo labels (He et al., 2021). Consequently, class-balanced approaches were proposed to increase the proportion of pseudo labels among difficult and rare classes (Zou et al., 2018), or instance-level adaptive thresholds that can be iteratively adjusted to match the prediction confidence present in input data (Mei et al., 2020). In the field of semantic segmentation for EO data, pseudo labels have been explored in generic land cover mapping tasks (Chen et al., 2022; Liu et al., 2023, 2022; Wu and Prasad, 2018), with most approaches conducting "online" pseudo label selection reliant on source data. For field delineation in smallholder agriculture, however, pre-trained models are scarce, and source data is oftentimes not readily available due to the reliance on commercial VHR imagery with prohibitive licensing agreements. Moreover, commonly used pseudo label selection workflows at the pixel level are unsuitable for field delineation, as many state-of-the-art field delineation architectures require semantically complete field instances (Cai et al., 2023; Li et al., 2023; Long et al., 2022; Waldner and Diakogiannis, 2020).

Exploring efficient tools for domain adaptation in source-free settings by facilitating the generation of high-quality instance-level labels at scale is a crucial research avenue to leverage the potential of artificial intelligence in smallholder systems, in particular in Sub-Saharan Africa

(Nakalembe and Kerner, 2023). We here explored the value of pseudo labels for overcoming domain gaps for field delineation in source-free settings. We used a pre-trained FracTAL ResUNet which has been previously deployed for field delineation in smallholder settings in India (Wang et al., 2022). We conducted closed-set domain adaptation with geographic, temporal, and sensor-specific domain shifts, where the size, shape, and appearance of fields in the target region (Mozambique) differed substantially from the source region (India). We used VHR RGB imagery (~0.6 m in target, 1.5 m in source) acquired from various airborne and satellite sensors, with different viewing angles, illumination conditions, and acquired during different times of the year. We tested the generation of semantically complete pseudo labels using multiple instance-level confidence thresholds and compared them with human-annotated labels to assess their spatial agreement, size, and site-level quantities across seasons. We assessed the value of each respective set of pseudo labels for overcoming domain gaps in two downstream applications, namely field delineation, and site-level field size estimation. For this, we fine-tuned the pre-trained field delineation model with each set of pseudo labels and assessed object-level spatial agreement and site-level field size estimates for our test sites. Our objectives were to: (1) create semantically complete pseudo labels for field delineation which are similar to human annotated labels, (2) assess the performance gains achieved by fine-tuning with pseudo labels relative to the performance gains obtained by using human labels for fine-tuning, and (3) explore the added value of pseudo labels as a complement to human labels in label-scarce settings.

# Methods

*Study area*

Our study area covers four provinces (Cabo Delgado, Nampula, Niassa, Zambezia) in the North of Mozambique, comprising an area of ~380,000 km² (Figure 1A). The farming system of the region is dominated by semi-subsistence smallholder farming, thus partly cultivating for subsistence while also engaging in commercial activities on local markets, and limited access to capital and a high reliance on manual labor (Figure 1B-D). Agricultural production is conducted with low inputs, such as fertilizer, pesticides, or herbicides, and access to mechanized equipment is rare. Key crops in the region are maize, beans, sorghum, cassava, soy, or non-food crops such as tobacco and cotton, as well as tree crops such as cashew and macadamia (Ministério da Agricultura e Desenvolvimento Rural, 2017).

Fields in the North of Mozambique are very small overall (Lesiv et al., 2019), but spatially explicit and nuanced data on the distribution of fields is not available. Based on our reference data, we estimated a range of field size between 0.001 ha and 1.4 ha, with a median of 0.06 ha. More than 99.8% of the fields in our reference data are smaller than one hectare, but substantial local-level variation (standard deviation = 0.16 ha) can be observed. Medium-scale farms, defined as those operating on land holdings of 5 – 100 ha, comprising fields of around 0.5 - 1 ha, are emerging in many parts of Sub-Saharan Africa (Jayne et al., 2016), including Mozambique (Baumert et al., 2019). Large-scale land acquisitions in the early 2000s led to the emergence of large fields covering more than one hectare, up to very large fields of more than 50, up to one hundred hectares, although they remain rare in relative terms (Bey et al., 2020).

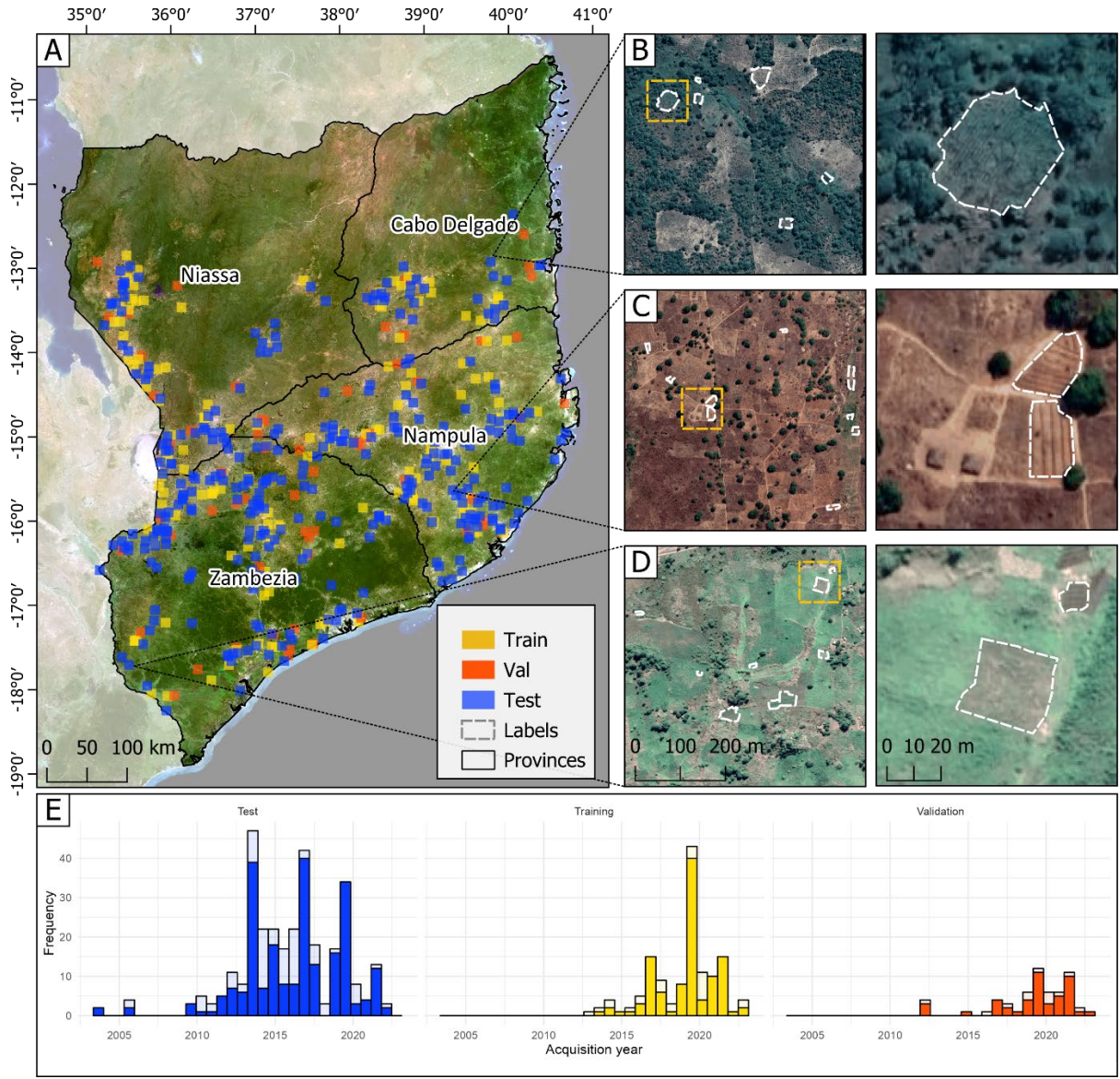

Figure 1: Study region in Northern Mozambique (A) with site locations for training (yellow), validation (red), and test split (blue). Image data sources: Panel A: PlanetScope mosaic of May 2023, provided through the NICFI data program (Planet Labs Inc., 2023). Panels B-D illustrate example sites with sparse labels and zoom ins. Image data sources: Google Earth Pro VHR imagery (©2023 Maxar Technologies). Panel E shows histogram of image acquisition years in each split, separated by dry season (filled bars) and wet season (hollow bars).

*Data*

*Very-high resolution imagery*

We used VHR image data contained in Google Earth Pro™. The image catalog of Google Earth contains imagery from Maxar Technologies, including WorldView 2 and 3 and GeoEye 1 imagery, as well as imagery acquired during airborne campaigns, all of which have a sub-meter spatial resolution. For site selection, we developed a stratified random sampling scheme to sample from regions with actively used cropland. To identify these regions, we used an existing map of active and fallow cropland for the growing season of September 2020 through August 2021 (Rufin et al., 2022). We aggregated the map to a 1 ha grid and calculated the proportions of active cropland. We sampled 1,000 sites from regions mapped as containing at least 50% of active cropland within a one-hectare grid cell. We defined a site extent of 600 by 600 meters, or 36 ha, in order to assure that a sufficient number of fields can be delineated, even in regions with comparatively large field size. The selected sites were screened for VHR image quality and acceptable visibility of at least five fields, resulting in 513 sample sites (see Figure 1A). We requested Google Earth™ imagery through the Google Maps API in R, resampled the RGB imagery to ~0.6 m per pixel by extracting 1024 x 1024-pixel images for rectangular sites of 600 x 600 m and stored them in GeoTIFF format. Image acquisition years varied between 2003 and 2022 (Figure 1E). Approximately 50% of the images were acquired in 2016 or later, and 2019 was the most frequent year for image acquisition. Over 75% of the images were acquired during the dry season between June and November.

*Labels*

Collecting field delineation labels in heterogeneous smallholder landscapes is challenging due to high levels of fragmentation, the presence of very small fields, high intra-field heterogeneity in crop cover, and a high reliance on manual labor, resulting in irregular field boundaries, which partly exert low contrast and are thus not clearly visible. Furthermore, collecting fully annotated labels with high confidence boundaries is in some cases hardly possible for regions where markers of field delineations are not distinct (Estes et al., 2022). In our study region, extensive cultivation in recently cleared forests represents one example of hard-to-interpret field delineations (Figure 1B). In such contexts, the use of weak supervision adds another benefit as it allows including field delineations from complex regions, where fully annotated labels can either not be created, or would imply sacrificing label quality by including uncertain field delineations.

We followed the routine described in Wang et al. (2022) and tasked human annotators to collect sparse labels (i.e. at least five fields per site). We tasked the interpreters to collect only fields containing non-tree crops by systematically excluding tree crop plantations from our data. While individual trees in the field interior were included in our labels, trees overlapping with the field boundaries were avoided and the tree canopy was considered as the field boundary for completing the labels. All field delineations underwent an iterative quality assessment, where 18% of the initial field delineations and 7% of the field delineations in a second iteration were discarded. We compiled a total set of 3,711 field delineations. We split the resulting 513 labeled sites into 200 training and 313 test sites, comprising 1,512, and 2,199 fields, respectively. We determined the number of labels required for training based on the ablation experiments in Wang et al. (2022), which indicated that 500 – 1,000 field labels were sufficient for fine-tuning a model to reach optimal performance. All labels used in this study are made publicly available for download via Zenodo: https://doi.org/XX.XXXX/zenodo.XXXXXXX.

*Model architecture*

We based our experiments on the DECODE framework, comprising a FracTAL ResUNet architecture for dense predictions of field extent, field boundaries, and within-field distance to the next boundary, which are used to generate semantically complete field instances using hierarchical watershed segmentation. The methods are described in detail in Waldner and Diakogiannis (2020), and Waldner et al. (2021). The FracTAL ResUNet resembles the typical UNet encoder-decoder architecture and features residual blocks with a built-in attention mechanism. The multiple model heads are designed for multi-task predictions on crop field extent, field boundaries, and the within-field distance to the nearest field boundary. The model was trained with a fractal Tanimoto loss to evaluate all tasks simultaneously. The FracTAL ResUNet outperforms other state-of-the-art model architectures in the task of field delineation (Tetteh et al., 2023), and has been shown to generalize well across regions (Waldner et al., 2021).

The DECODE framework was used by Wang et al. (2022), who reported high performance in smallholder agricultural regions of India using multi-spectral pan-sharpened SPOT6/7 data at 1.5 m spatial resolution. For model training, Wang et al. (2022) processed the four-channel input data (RGB and near-infrared) to a three-channel false-color representation with channel 1 as the average of the red and green bands, band 2 as the average of the red and NIR bands, and band 3 as the average of the green and blue bands. The authors conducted pre-training using publicly available field delineations from the land parcel information system in France. They fine-tuned the model for field delineation in India using sparse labels created by human annotators across 2,000 sites. Due to the sparse annotation procedure, only five fields per site were included in the training, which allowed for a broader geographic coverage and, thus, a higher diversity of fields in the training data.

*Pseudo label selection*

We tested the potential of using a pre-trained model for producing semantically complete field instances that can be used as training and validation labels to fine-tune models for Mozambique, a structurally different and more complex target region than France and India.

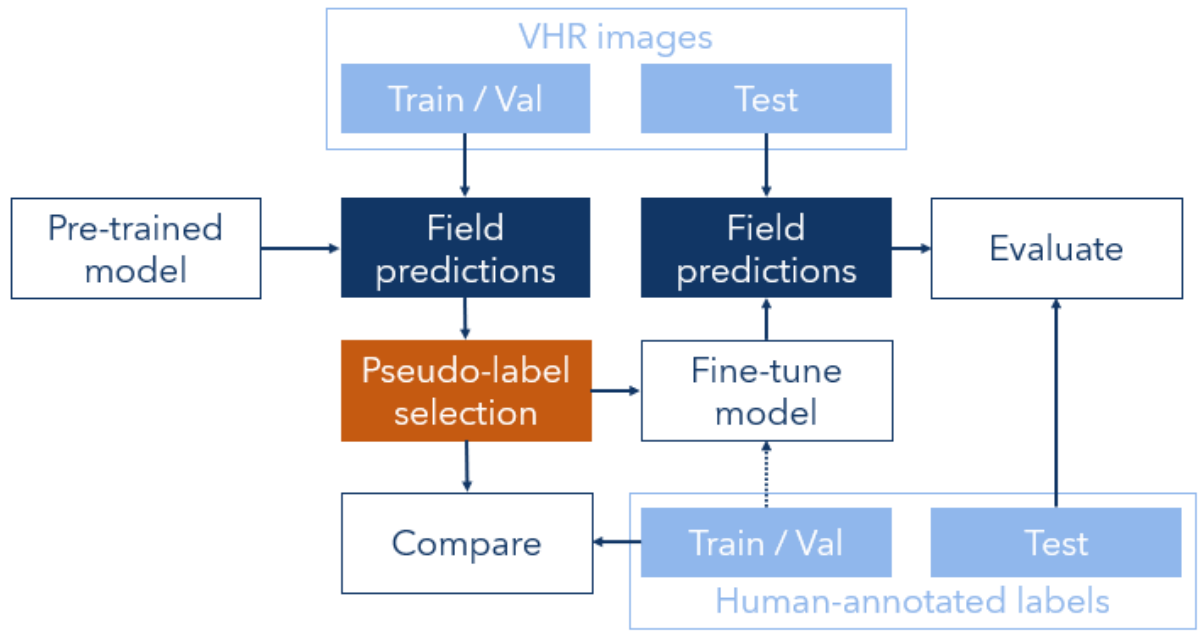

Figure 2: Workflow of this study.

We assessed the performance of the publicly available, pre-trained FracTAL ResUNet by Wang et al. (2022) to predict fields in Mozambique, which can be used as pseudo labels for fine-tuning the model. Given the assumption that the basic properties of field appearance across geographies are similar, we reason that the pre-trained model can produce accurate field delineations in the known feature space without fine-tuning. We used the model to create predictions of cropland and boundary probabilities for our 200 training sites, and performed instance generation from the predictions based on hierarchical watershed segmentation (Waldner et al., 2021). We kept the hyperparameters of the watershed segmentation constant at boundary threshold = 0.2 and extent threshold = 0.4 following Wang et al. (2022) for better comparison across model runs while acknowledging that optimizing the hyperparameters based on validation labels may further improve performance.

We then calculated a set of instance-level scores using the model output activations (sigmoid-scaled in the case of field boundary predictions) for both cropland field extent and field boundaries. For pseudo label selection, we derived the semantic confidence $SemC_i$, for each instance $i$ comprising $n$ pixels as the median of the pixel-level field extent probabilities $p_{ext}$:

$$SemC_i = P_{50}(p_{ext})$$

We derived instance confidence $InsC_i$ relating to the median boundary probabilities $p_{bnd}$ in all boundary pixels of instance $i$ as:

$$InsC_i = P_{50}(p_{bnd})$$

Lastly, we calculated the size of each instance as the number of pixels $n$ comprising the instance:

$$Size_i = n$$

We first tested selection based on thresholding $SemC_i$ using absolute thresholds $T_{SemC}$, so that a pseudo label was selected if:

$$SemC_i > T_{SemC}$$

We tested four absolute probability thresholds $T_{SemC} = \{0.925, 0.950, 0.975, 0.990\}$.

The absolute thresholds constrain pseudo label selection to the overall most certain predictions, which reduces the likelihood of including erroneous pseudo labels, but may limit pseudo label selection to images and settings with comparatively high confidence and may thus reduce the diversity of pseudo labels. To increase diversity of pseudo labels, we tested adaptive thresholds based on the site-level distribution of confidence scores. Thresholds were derived using percentiles of all instances-level scores per site $P_j(SemC_N)$ with $j$ denoting the respective percentile, and $N$ being all instances within the site. We considered $j = 99$, $j = 98$, and $j = 95$, thus selecting the 1%, and 2%, most confident predictions, respectively. When targeting a pseudo label selection based on $P_j(SemC_N)$ and $P_j(InsC_N)$ simultaneously, we defined $j = 95$, i.e., selecting the top 5% from each metric, because instances with both high semantic and high instance confidence were rare:

$$SemC_i > P_{95}(SemC_N) \ \& \ InsC_i > P_{95}(InsC_N)$$

Moreover, we considered a minimum size criterion to avoid the selection of small fragments consisting of a few confident pixels, such as individual trees which were apparent in the predictions. We defined a minimum size of $T_{Size}$ = 180 m² (i.e. 500 pixels) which corresponds to the extent of a large tree canopy, as often present on or nearby fields in the region. We tested the size criteria in combination with $P_{99}(SemC_{image})$ only. Following this logic, an instance was selected from the pool of pseudo label candidates if

$$SemC_i > P_{99}(SemC_N) \ \& \ Size_i > T_{Size}$$

Following this procedure, we compiled a total of eight sets of pseudo labels, with four selected based on absolute and four based on adaptive thresholds. We then compared human and machine-generated labels in terms of label quantity, area properties, and spatial agreement. Besides testing the use of pseudo labels for fully replacing human-annotated data, we also tested the value of pseudo labels as a complement to human labels. For this, we generated pseudo labels for an additional set of images (n=400) following the percentile-based routine described above and complemented the pseudo labels with human-annotated data for the 200 training sites, thereby providing 600 training sites to the model. The code used to produce pseudo labels will be made publicly accessible via GitHub (https://github.com/philipperufin/pseudo-fields/).

*Non-cropland labels*

One caveat of the weak supervision strategy employed by Wang et al. (2022) is that the model did not learn to discriminate cropland from non-cropland, as reference data for non-cropland classes were not included during training. Reliable attribution of instances to cropland or other land cover or land use types thus requires a detailed, timely, and accurate cropland mask or additional updating of the non-cropland classes. We leveraged the uncertain predictions of the pre-trained model to add patches of non-cropland samples to training. We complemented non-cropland pseudo labels for the absolute thresholds by selecting instances if:

$$\text{SemC}_i < T_{\text{SemC}} = 0.75,$$

For the adaptive strategy, we defined $j = 10$, selecting non-cropland labels if:

$$SemC_i < \text{P}_{10}(SemC_N),$$

For the combined selection based on semantic and instance confidence, we selected an instance of non-cropland if:

$$\text{SemC}_i < \text{P}_{25}(SemC_N) \text{ \& } \text{InsC}_i < \text{P}_{25}(InsC_N).$$

We did not account for a minimum instance size during negative pseudo label selection due to the complexity of the landscapes with highly variable instance sizes depending on the underlying non-cropland land cover type. An example for field and non-cropland pseudo label selection is illustrated in Figure 4.

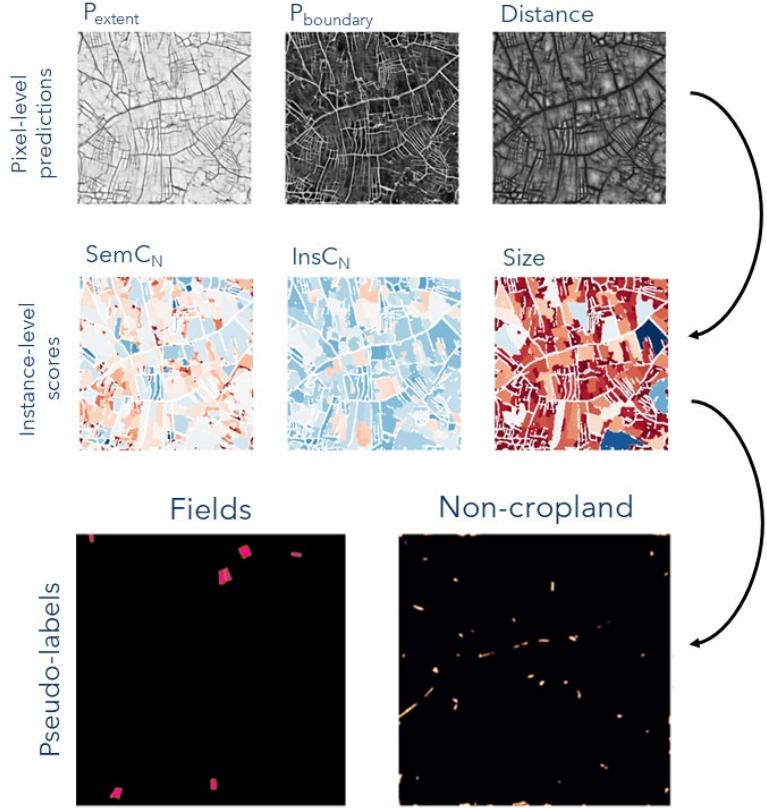

Figure 3: Example for pseudo label generation using $SemC_i > P_{99}(SemC_N)$ for field labels and $SemC_i < P_{10}(SemC_N)$ for non-cropland labels. Predictions of the input image (top row) are used to generate instances, for which scores are computed (middle row). Pseudo labels for fields and non-cropland are selected from all instances based on the respective selection strategy, here $SemC_i < P_{10}(SemC_N)$ (bottom row).

*Model fine-tuning*

We used the training data obtained from human annotators and the five sets of machine-generated training data to update the pre-trained FracTAL ResUNet in ten experiments, namely 1) fine-tuning using the sparse labels generated from human annotators, 2) – 5) fine-tuning using pseudo labels with $SemC_i > T_{SemC}$ with $T_{SemC}$ = 0.925, 0.950, 0.975, 0.990, 6) $SemC_i > P_{99}(SemC_N)$, vii) $SemC_i > P_{99}(SemC_N)$ & $Size_i > 180m^2$, viii) $SemC_i > P_{98}(SemC_N)$, ix) $SemC_i > P_{95}(SemC_N)$ & $InsC_i > P_{95}(InsC_N)$, and x) human labels combined with pseudo labels with $SemC_i > P_{99}(SemC_N)$ for 400 additional sites.

From the pseudo labels, we generated multi-task labels representing field boundaries, field extent, and within-field distance to the nearest boundary using Python's OpenCV module (Bradski, 2000). For model updating, we cropped the 1024 x 1024-pixel images into sixteen 256 x 256-pixel image chips, resulting in 2,240 image chips. We removed chips without any field labels. This resulted in a varying number of image chips for the different fine-tuning experiments. The 200 training sites were split into 70% (n = 140) for training and 30% (n = 60) for validation. We ran the fine-tuning for 100 epochs with a batch size of 4 and a start learning rate of 0.0001 using Adam optimizer in all experiments to obtain comparable insights on training time and performance. We selected the best model based on the maximum Matthew´s correlation coefficient in the validation split and registered the number of epochs used for training.

*Performance evaluation*

*Field-level spatial agreement*

For each resulting model, we predicted field boundaries for the 313 test sites and created field instances based on watershed segmentation as described in Wang et al. (2022). We vectorized the predicted instances using GDAL (GDAL/OGR contributors, 2023), matched predicted fields with reference fields using the centroid of the reference fields, and calculated the Intersection over Union (IoU), Precision, and Recall of each field instance using the segmetrics package in R (Simoes et al., 2022). We aggregated all field-level IoU scores using mean (mIoU), median (medIoU), the share of IoU scores above 0.5 ($IoU_{50}$) and 0.8 ($IoU_{80}$) conducted in-depth error diagnoses by assessing instance-level IoU scores by field size, the season of image acquisition, and province. The resulting insights on the error distribution were used to test for performance declines in particular subsets of the target domain when considering the different sets of pseudo labels. We quantified pseudo label performance gains as percent relative to the performance obtained with human labels to assess to what degree domain adaptation can be achieved without the need for any human labels:

$$relative\ gain = \left(\frac{score_{pseudo} - score_{baseline}}{score_{human} - score_{baseline}}\right) * 100$$

*Site-level field size estimation*

We further evaluated the suitability of the respective models for estimating field size at the site level. The assumption here was that while object-level performance scores may correlate with field size estimation performance, the latter relaxes the constraints on the spatial alignment of the individual field delineations while maintaining focus on the overall distribution of field size present in the predictions. We computed the root mean squared error (RMSE), mean absolute error (MAE), and mean error (ME) at the site level. We then averaged the site-level scores across the test sites (*n* = 313) to obtain mean RMSE (mRMSE), mean MAE (mMAE), and mean bias (mME). We further investigated the spatial distribution of the RMSE scores across provinces and calculated global Moran´s I to test for spatial autocorrelation in the model errors. Additionally, we fitted linear regression models between observed and predicted field sizes and recorded the goodness-of-fit as $R^2$, slope, and intercept of the regression. Similar to the object-level assessment, we quantified performance gains from pseudo labels relative to performance gains obtained from human labels as percent of the performance increase (or error reductions, respectively) of using human labels against the pre-trained model.

*Non-cropland identification*

We evaluated the discrimination of cropland against other types of land cover and land use that was only trained during fine-tuning based on non-cropland pseudo labels. While we acknowledge that a robust evaluation of the separation between cropland and non-cropland is ideally conducted on fully annotated labels, the landscape complexity did not allow for compiling fully segmented labels on non-cropland. We therefore compiled sparsely annotated non-cropland labels for the test sites, which we could derive from an existing map product indicating the presence of cropland at approximately 5 m spatial resolution for the growing season 2020/2021 (Rufin et al., 2022). We considered all regions mapped as open woodlands, closed woodlands, non-vegetated surfaces, and water with an area comprising at least four pixels (or ~91 m²) as candidates for evaluating the ability of the model to separate non-cropland. Due to the mismatches between spatial resolutions and acquisition dates for most images in the test dataset, we manually screened all candidate regions to remove all non-cropland patches that were not representing the non-cropland domain accurately. We thereby compiled a final set of 2,143 reference data patches representing the non-cropland domain in our study region. We evaluated the non-cropland identification in the region using pixel-level overall accuracy, Precision, Recall, and F1-score at the image level and as average across all images in the test set. Moreover, we conducted an in-depth qualitative evaluation of the non-cropland identification to support the empirical data with qualitative insights.

# Results

*Pseudo labels for field delineation*

We compared the quantities and area properties of the pseudo labels with human labels for the 200 training sites (Table 1, Figure 4). When using absolute thresholds, the number of selected instances in the pseudo labels as well as the average number of labels per site decreased with increasing confidence thresholds, with values ranging from 3,472 field instances for the $T_{SemC} = 0.925$, to 267 field instances for $T_{SemC} = 0.990$, against 1,517 field instances in our human-annotated data. A minimum of 5 fields per site was obtained at 163 sites for the human-annotated data, and between 167 ($T_{SemC} = 0.925$) down to only 12 ($T_{SemC} = 0.990$) sites for the pseudo labels. Generally, the spread in the number of fields per site was higher for the pseudo labels than the human-annotated data, where in some cases, up to 70 fields ($T_{SemC} = 0.925$) or 38 fields ($T_{SemC} = 0.990$) were generated. For several sites, no pseudo labels could be generated, with the number of sites without pseudo labels ranging from 5 ($T_{SemC} = 0.925$) up to 133 when using the most conservative threshold ($T_{SemC} = 0.990$). For the adaptive thresholds, we observed a lower variation in the number of selected fields compared to the absolute thresholds. Selecting the top 1% predictions ($P_{99}(SemC_N)$) yielded 766 fields, or 695 when including with the minimum size criterion. Relaxing the threshold to $P_{98}(SemC_N)$ yielded 1,334 field instances, while the selection based on $P_{95}(SemC_N)$ and $P_{95}(InsC_N)$ combined yielded only 342 field instances (Table 1, Figure 4).

The area properties of both the human and pseudo labels selected with absolute thresholds were more similar when applying conservative thresholds. Discrepancies increased when relaxing the threshold, and the number of large fields increased substantially when using lower confidence thresholds. Mean and median field sizes in the human labels (0.120 ha, and

0.060 ha) and $T_{SemC} = 0.990$ pseudo labels (0.120 ha, and 0.080 ha) agreed well (Table 1). The standard deviation for the human-annotated labels was slightly higher (0.164 ha) as compared to the pseudo labels (0.114 ha). Using the adaptive thresholds, the mean area was closest to the human labels with $P_{95}(SemC_N)$ and $P_{95}(InsC_N)$ combined, followed by $P_{99}(SemC_N)$. We found that using $P_{95}(SemC_N)$ and $P_{95}(InsC_N)$ yielded a range of field sizes comparable to our human labels but with an overrepresentation of small fields with a size of approximately 0.02 ha and less (Figure 4).

Table 1: Descriptive statistics of human-annotated and machine-generated labels.

|  | Human labels | Absolute thresholds | | | | Adaptive thresholds | | | |
| --- | --- | --- | --- | --- | --- | --- | --- | --- | --- |
|  |  | $T_{SemC} =$ 0.925 | $T_{SemC} =$ 0.950 | $T_{SemC} =$ 0.975 | $T_{SemC} =$ 0.990 | $P_{99}$ $(SemC_N)$ | $P_{99}(SemC_N)$ & Size | $P_{98}$ $(SemC_N)$ | $P_{95}(SemC_N)$ & $P_{95}(InsC_N)$ |
| Total N fields | 1,517 | 3,472 | 2,406 | 935 | 267 | 766 | 695 | 1,334 | 342 |
| Mean N fields | 7.74 | 17.36 | 12.03 | 4.68 | 1.34 | 3.83 | 3.475 | 6.67 | 1.71 |
| Max N fields | 24 | 70 | 61 | 64 | 38 | 13 | 10 | 19 | 12 |
| N sites ≥ 5 fields | 163 | 167 | 137 | 55 | 12 | 70 | 54 | 152 | 15 |
| N sites 0 fields | 4 | 5 | 18 | 71 | 133 | 3 | 3 | 0 | 68 |
| Mean (ha) | 0.1187 | 0.6856 | 0.4052 | 0.1727 | 0.1172 | 0.1285 | 0.1326 | 0.2114 | 0.1161 |
| Median (ha) | 0.0600 | 0.1162 | 0.1113 | 0.0950 | 0.0806 | 0.0741 | 0.0771 | 0.0797 | 0.0594 |
| SD (ha) | 0.1642 | 2.7494 | 1.5031 | 0.3466 | 0.1144 | 0.2149 | 0.2176 | 1.2423 | 0.2305 |

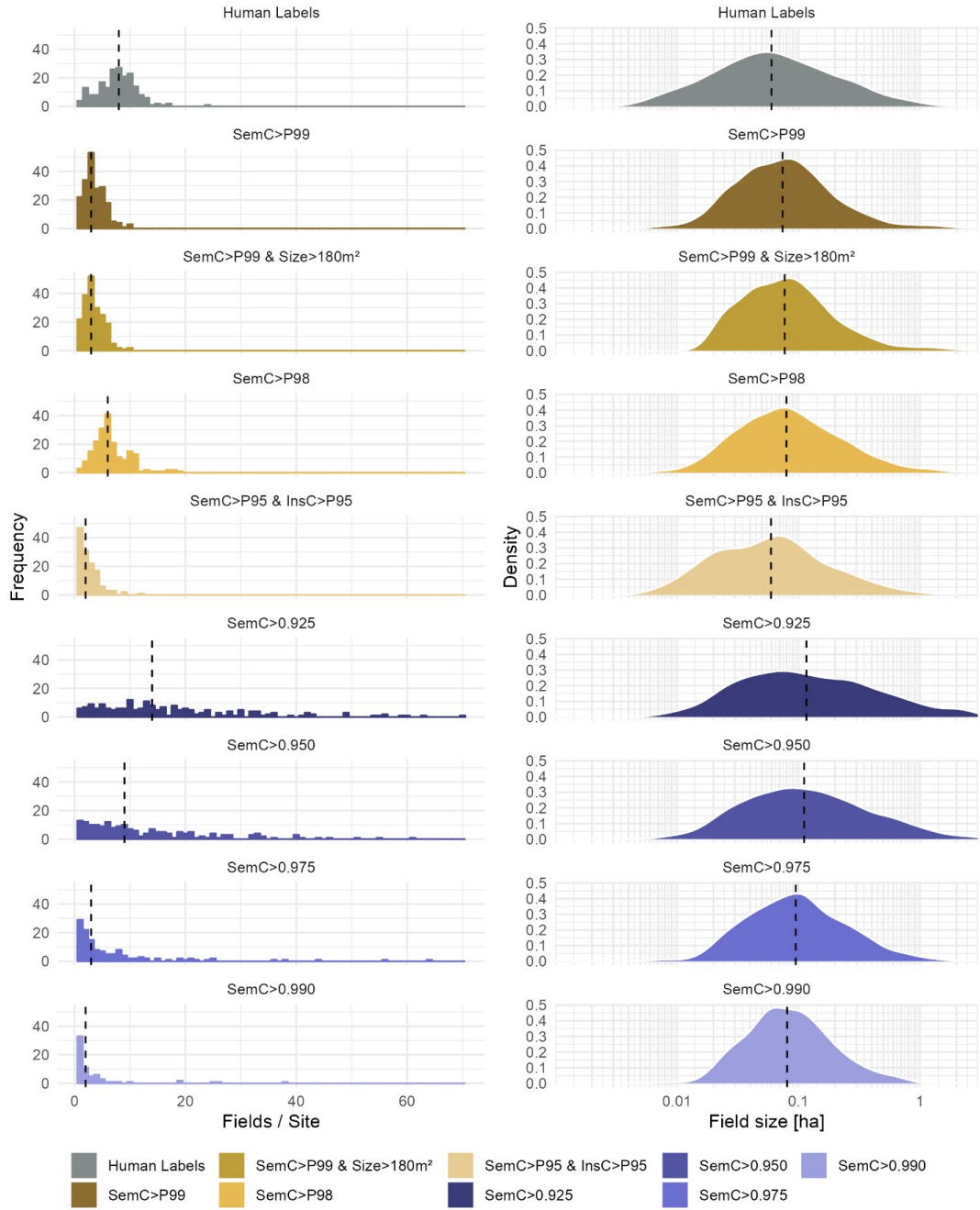

Figure 4: Number of fields per site (left column) and density curves of field size in hectares (right column) of human labels (top row) and pseudo labels with varying thresholds (rows 2-9). Note that x-axis of field size is log-scaled.

We found high spatial agreement between the human-annotated labels and the pseudo labels with increasing agreement at higher confidence thresholds (Table 2). The pseudo labels selected at $T_{\text{SemC}}$ = 0.990, reached a mIoU of 0.802, medIoU of 0.868, IoU$_{50}$ of 0.914, IoU$_{80}$ of 0.743, and a high Precision of 0.871 and Recall of 0.904. Importantly, Precision and Recall were also balanced in this case.

Table 2: Performance metrics comparing pseudo labels with human-annotated labels for training sites.

|  | $T_{SemC}$ = 0.925 | $T_{SemC}$ = 0.950 | $T_{SemC}$ = 0.975 | $T_{SemC}$ = 0.990 | $P_{99}$ ($SemC_N$) | $P_{99}(SemC_N)$ & Size | $P_{98}$ ($SemC_N$) | $P_{95}(SemC_N)$ & $P_{95}(InsC_N)$ |
|---|---|---|---|---|---|---|---|---|
| mIoU | 0.629 | 0.663 | 0.743 | 0.802 | 0.741 | 0.741 | 0.735 | 0.732 |
| medIoU | 0.741 | 0.777 | 0.821 | 0.868 | 0.804 | 0.807 | 0.804 | 0.811 |
| $IoU_{50}$ | 0.729 | 0.773 | 0.857 | 0.914 | 0.855 | 0.853 | 0.847 | 0.845 |
| $IoU_{80}$ | 0.396 | 0.439 | 0.571 | 0.743 | 0.503 | 0.509 | 0.510 | 0.563 |
| Dice | 0.561 | 0.554 | 0.830 | 0.887 | 0.828 | 0.827 | 0.816 | 0.797 |
| Precision | 0.422 | 0.406 | 0.769 | 0.871 | 0.779 | 0.779 | 0.763 | 0.709 |
| Recall | 0.838 | 0.868 | 0.902 | 0.904 | 0.882 | 0.882 | 0.876 | 0.911 |

We observed slight differences in the number of pseudo labels per site between the dry season, here defined as June through November, and the wet season, lasting from December through May (Figure 5). Differences between seasons were statistically insignificant (two-tailed t-test) for the human labels and absolute confidence thresholds (p>0.1), but significant differences (p<0.05) were observed for three adaptive selection strategies, with a higher average number of labels for wet season observations as compared to dry season observations for $P_{99}(SemC_N)$ (dry season: 3.4, wet season: 4.4, p=0.015), $P_{98}(SemC_N)$ (dry season: 6.4, wet season: 8.0, p=0.022), and $P_{95}(InsC_N)$ and $P_{95}(InsC_N)$ (dry season: 1.6, wet season: 2.5, p=0.020).

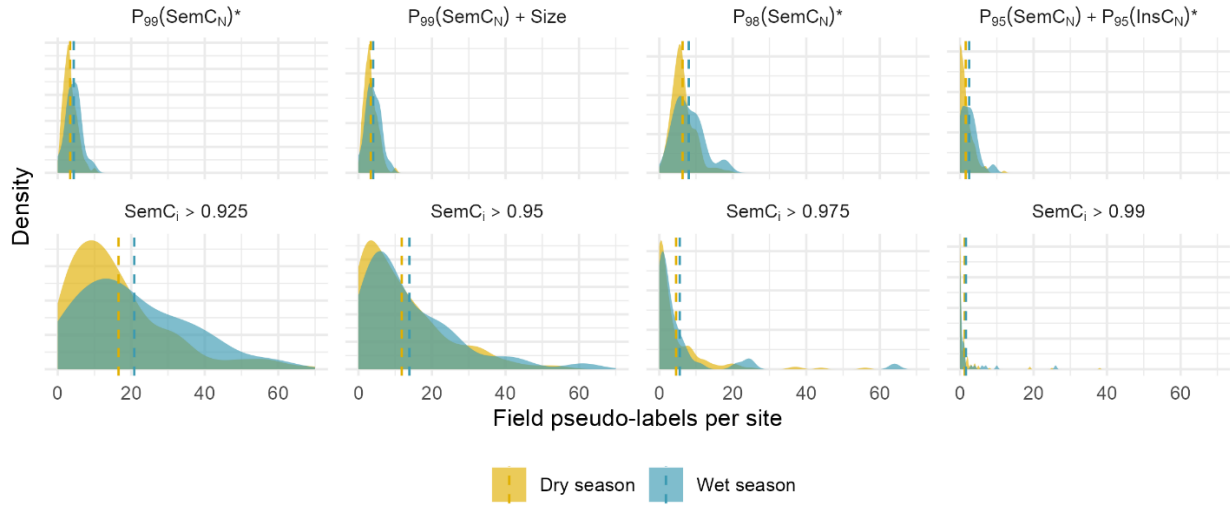

Figure 5: Number of machine-generated reference fields per site by season of image acquisition. Pseudo labels with significant (p<0.05) differences between dry season (yellow) and wet season (blue) are marked with asterisk.

*Fine-tuning & learning curves*

The learning curves for all fine-tuning runs indicate stable learning behavior. Maximum MCC scores on the validation split were reached in 20 ($T_{SemC}$ = 0.990) to 57 ($P_{99}(SemC_N)$ & Size) epochs. A direct comparison of the absolute loss and metric values in the training and validation split was not possible, as they differed in terms of the number of image chips containing labels and the characteristics of the fields therein.

*Object-level performance*

We assessed object-level agreement between predictions of the fine-tuned models and our independent reference data using mIoU, IoU50, Precision (or User´s accuracy), and Recall (or Producer´s accuracy) (Figure 6). We used the pre-trained model as a baseline to assess relative performance improvements, while the model trained based on human labels represented the target. Fine-tuning with human labels increased performance by 0.052 (8%) for mIoU and 0.073 (10%) for $IoU_{50}$, respectively.

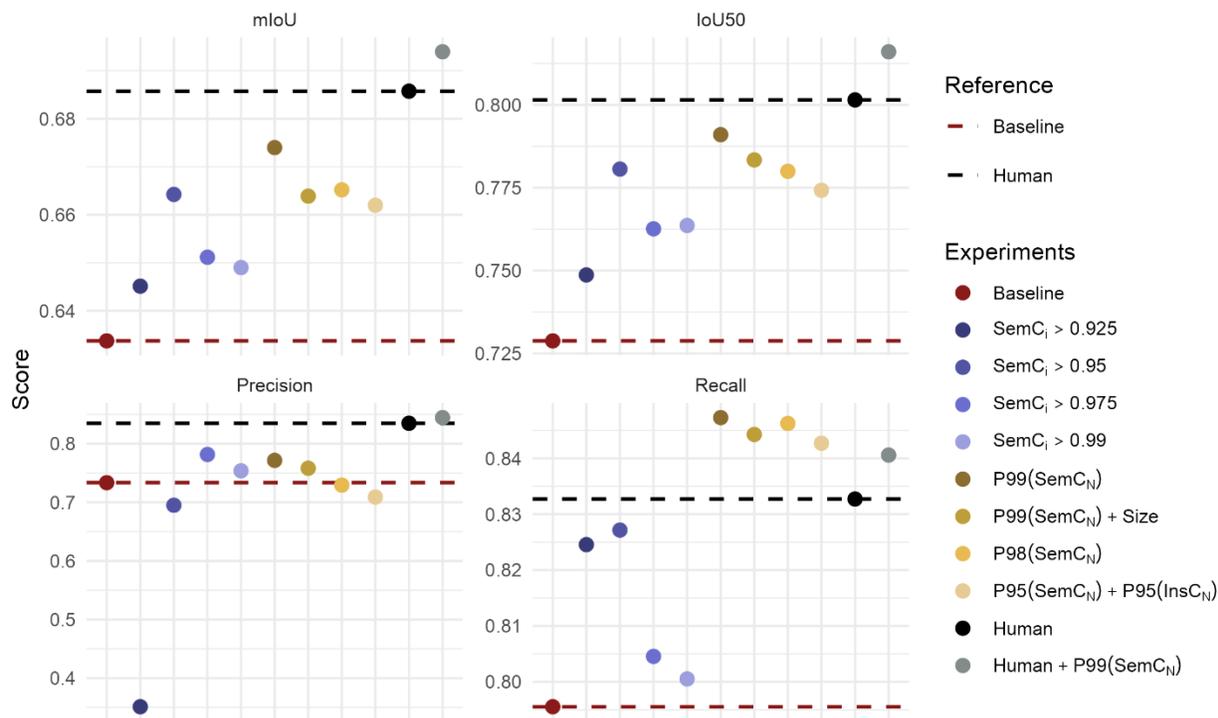

Figure 6: Comparison of mIoU, IoU50, Precision, and Recall across all experiments. Dashed lines represent baseline model (red) and model trained with human labels (black).

When using absolute thresholds, performance declined with lower threshold values. We found the lowest performance of all pseudo labels at $T_{SemC} = 0.925$, except for a substantial increase in Recall, which was, however, outweighed by a decline in Precision, indicating substantial undersegmentation. The high absolute thresholds $T_{SemC} = 0.975$ and $T_{SemC} = 0.990$ yielded performance gains across all metrics, although with moderate magnitude (e.g. 29% of mIoU and 48% of $IoU_{50}$ increases relative to performance with human labels).

Adaptive thresholds in pseudo label selection led to overall higher performance gains. The combination of $P95(SemC_N)$ and $P95(InsC_N)$ performed the worst, followed by the selection using $P98(SemC_N)$. $P99(SemC_N)$ outperformed all other approaches at the object level and consistently led to substantial performance increases. Further constraining pseudo label selection using size thresholds degraded the performance gains. Fine-tuning with pseudo labels based on $P99(SemC_N)$ allowed for achieving 77.4%, and 85.6% of the human label performance gains in mIoU, and $IoU_{50}$, while consistently reducing errors across field sizes and sub-regions.

Combining human-annotated data with pseudo labels for additional sites improved the performance of the model across all scores. At the object level, mIoU was 0.686 using human labels and 0.694 when complemented with pseudo labels, and $IoU_{50}$ increased from 0.801 to 0.816. In relative terms, this relates to a performance increase of 115.8% for mIoU and 120.0% for $IoU_{50}$ relative to the performance gains obtained from human labels.

In-depth error diagnoses reveal the distribution of performance gains across the observed range of field size and across different parts of the study region in more detail (Figure 7). IoU curves reveal that the overall distribution of IoU scores improved for all fine-tuning experiments relative to the baseline model. Assessing mIoU across the range of field size revealed the complexities with correctly identifying very fields smaller than 0.01 ha, which was apparent in all models. For fields above 0.01 ha, all fine-tuned models had substantially improved performance over the pre-trained baseline model. For fields above 0.05 ha, the models involving human labels outperformed the models trained exclusively with pseudo labels. Comparing human labels only against combined human and pseudo labels reveals similar performance across the range of field size, with the combined model slightly outperforming human labels only across the entire range. Assessing median model performance regionally, i.e., across provinces, further attested that combining pseudo labels with human labels performs at least similarly well, or better than human labels only. Using only pseudo labels was on par with using only human labels in Nampula province, and pseudo labels even outperformed human labels in Niassa province. Pseudo labels did not yield comparable performance in Cabo Delgado and Zambezia province. Differences in IoU between predictions in dry season versus wet season imagery were statistically insignificant (p > 0.1) in all model runs.

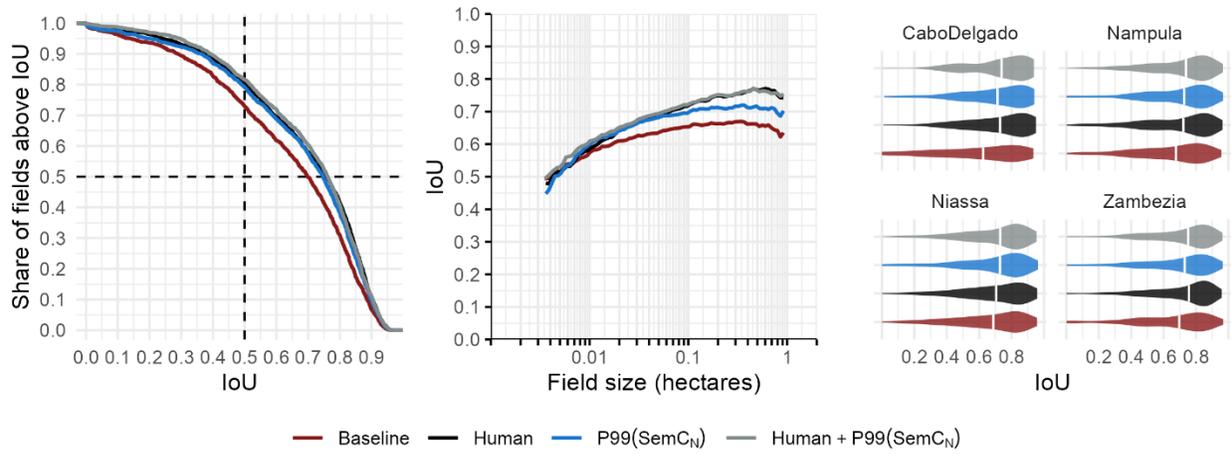

Figure 7: Comparison of for pre-trained baseline model (red), pseudo labels selected with $P99(SemC_N)$ (blue), human labels (black), and combined human labels and pseudo labels (azure). IoU curves (left), mIoU across field size (middle), and province-level IoU distribution with median in white (right).

A qualitative assessment of the field predictions (Figure 8) attested to the accurate field delineation results and how regional fine-tuning helps to overcome context-specific issues of under- (Figure 8A, C, E), or oversegmentation (Figure 8D). The pre-trained model performed well in some cases (Figure 8B, D), attesting to the hypothesis that accurate delineations can be created for individual (but not all) fields when transferring models across geographies. However, there were cases where neither the pre-trained nor the fine-tuned models produced optimal field delineations (Figure 8F).

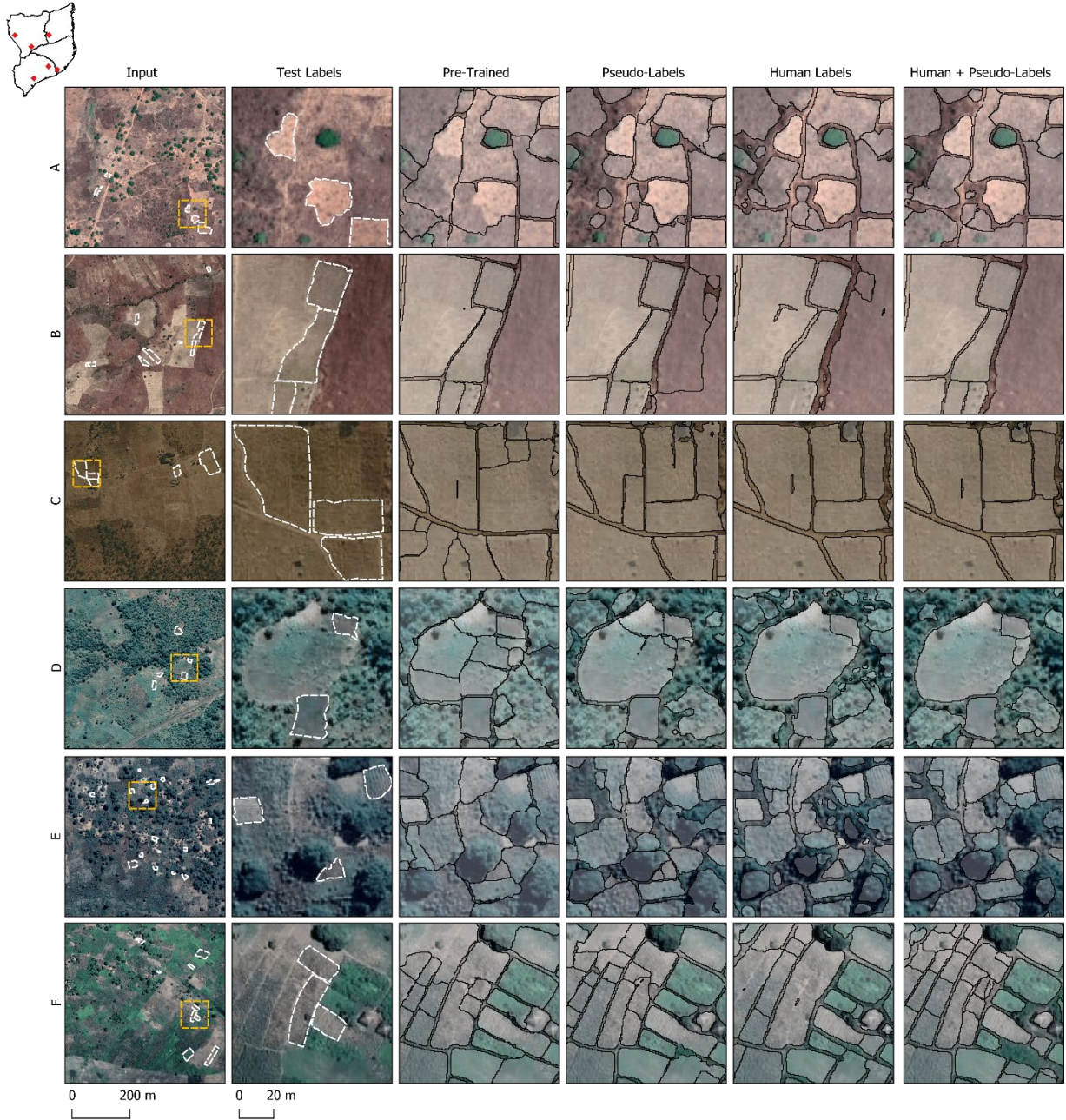

Figure 8: Selected test sites with Google Earth RGB images (©Maxar Technologies 2019) (left column, yellow dashed line is the area zoomed in to in columns 2-6), labels (second column), predictions (columns 3-6) using the pre-trained model, pseudo labels, human labels, and human labels combined with pseudo labels.

*Field size estimation performance*

We assessed site-level performance based on mean RMSE (mRMSE), median RMSE (P50RMSE) and mean MAE (mMAE) scores, as well as adjusted $R^2$ of observed against predicted field size (Figure 9). Compared to the baseline, fine-tuning with human labels reduced mRMSE by 0.032 ha, from 0.071 ha in the baseline to 0.040 ha, mMAE by 0.024 ha, from 0.050 ha in the baseline to 0.027 ha, and increased the R² of observed vs. predicted by 0.153, from 0.646 to 0.799, respectively.

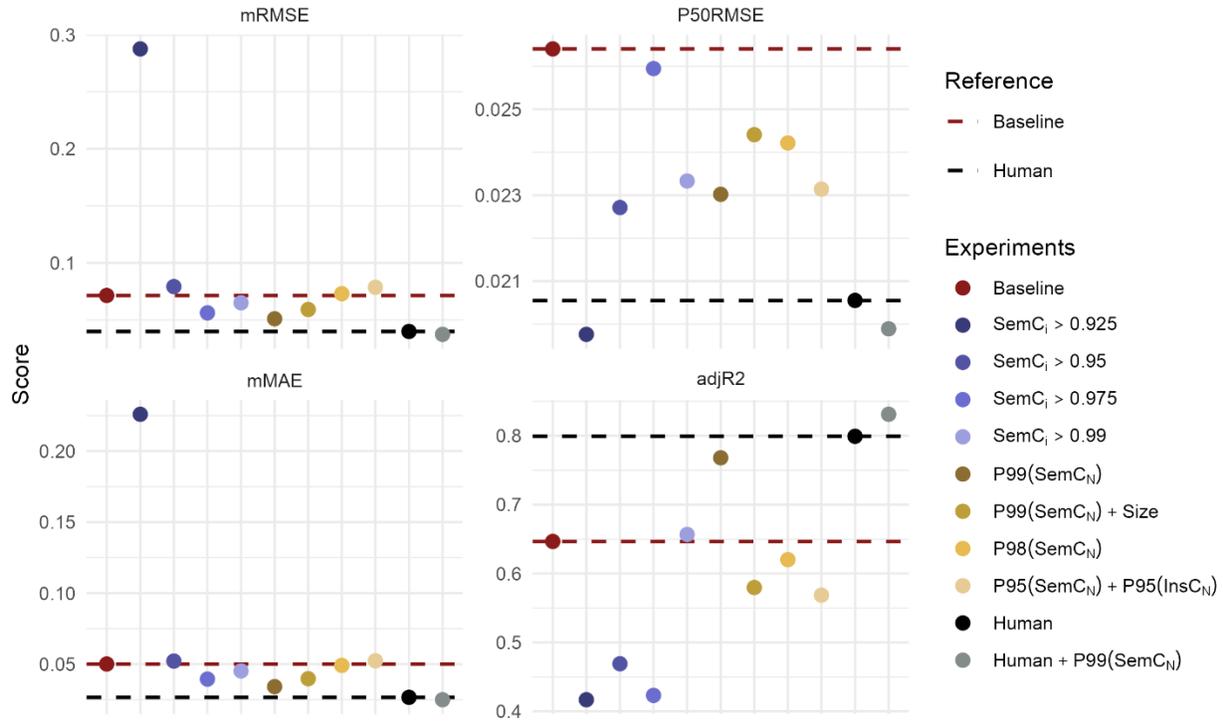

Figure 9: Comparison of mRMSE, P50RMSE, mMAE, and R² across experiments. Dashed lines represent baseline model (red) and model trained with human labels (black).

Similar to the object-level analyses, using liberal absolute thresholds ($T_{SemC}$ = 0.925 and $T_{SemC}$ = 0.950) led to a performance below baseline for mRMSE, mMAE, and R². More conservative thresholds ($T_{SemC}$ = 0.975 and $T_{SemC}$ = 0.990) yielded performance increases in mRMSE and mMAE, but only $T_{SemC}$ = 0.990 led to increases in R². Again, $P99(SemC_{image})$ consistently outperformed all other approaches. Fine-tuning with pseudo labels based on $P99(SemC_{image})$ achieved performance gains of 68.2%, 65.8%, and 68.8% for mRMSE, mMAE, and R² relative to human labels, respectively.

Combining human-annotated data with pseudo labels for additional sites improved the performance across all metrics. We observed an mRMSE decrease from 0.040 ha using only human labels to 0.037 ha when complementing with pseudo labels, and R² increased from 0.799 to 0.831, representing 108.1% and 120.9% of relative performance gains, respectively.

Error diagnoses revealed the high agreement between observed and predicted field size when using human labels or combined labels and how distributions of RMSE shift towards lower RMSE (Figure 10). Using human labels while consistently reducing median errors across the four sub-regions. Combining pseudo labels with human labels outperformed human labels only, except for Cabo Delgado province, where we noted a decrease in performance as compared to human labels only. To account for spatial autocorrelation in model errors, we assessed global Moran´s I statistics of the site-level RMSE scores. We did not find a significant (p<0.1) spatial autocorrelation in the field size residuals.

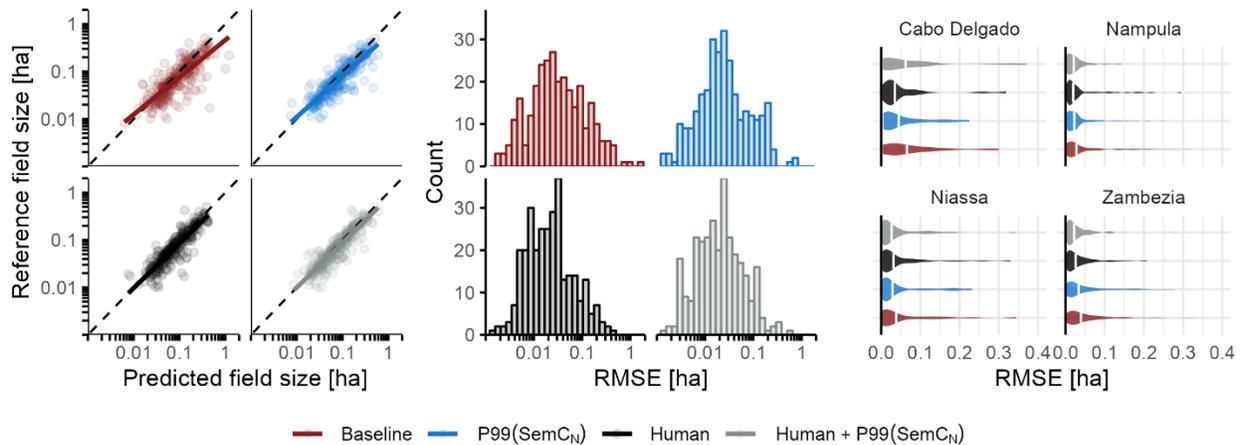

Figure 10: Comparison of for pre-trained baseline model (red), pseudo labels selected with $P99(SemC_N)$ (blue), human labels (black), and combined human labels and pseudo labels (azure). Plots show linear regression lines between reference field size and predicted field size (left), histogram of site-level RMSE scores (middle), and site-level RMSE across the four different provinces with median in white (right).

*Non-cropland identification performance*

The quantitative assessment of the non-cropland identification revealed a general tendency to underestimate non-cropland. While errors of commission were rare in the non-cropland domain, errors of omission were common. This resulted in high Precision (mean Precision = 0.822), and low Recall (mean Recall = 0.511) scores for the non-cropland identification (Figure 11). The mean accuracy (i.e. combining the cropland and non-cropland classes) across all test sites was 0.652, which is not satisfactory but a large step forward given that the initial model was not able to identify non-cropland at all, and given that the improvement has been achieved only by including non-cropland pseudo labels in fine-tuning.

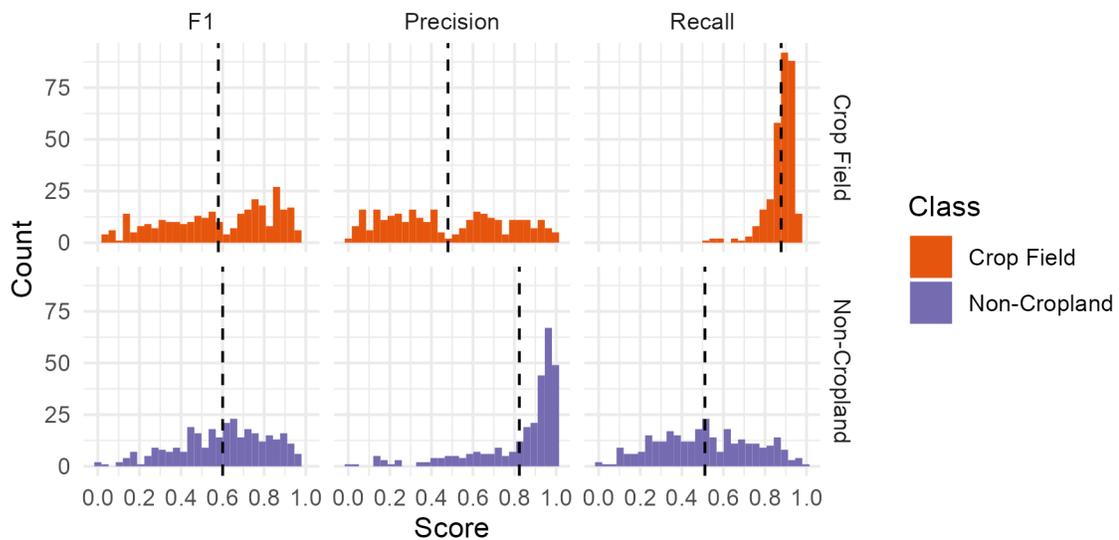

Figure 11: Quantitative evaluation of non-cropland discrimination

A qualitative evaluation of the field delineation and the non-cropland discrimination when using non-cropland pseudo labels confirmed the qualitative findings of lacking discriminative capacities regarding the non-cropland identification. Larger continuous patches of open or closed tree cover were correctly mapped as non-cropland in most cases, although smaller patches therein were partly wrongly mapped as fields (Figure 12A). In some cases, the improved recognition of non-cropland, relative to the pre-trained model, helped in discriminating isolated fields surrounded by non-cropland (Figure 12B). Individual trees were in many cases correctly identified as non-cropland (Figure 12C), but were falsely identified as fields in other cases, including for clusters of trees (Figure 12D) or large trees with unevenly shaped canopies (Figure 12E). Fallow lands were often correctly excluded from the cropland domain, although the discrimination was sensitive to the structure and texture and not consistent in all cases (Figure 12F).

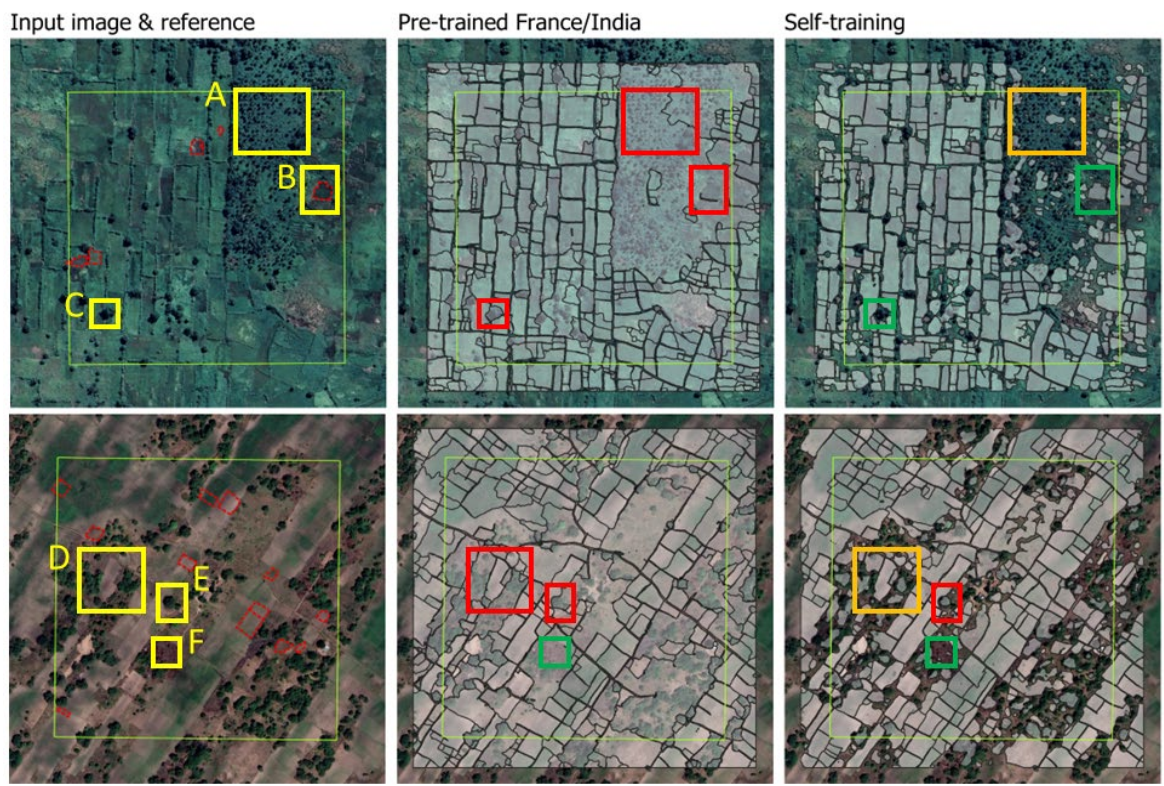

Figure 12: Example input imagery with human labels (first column), predictions of the baseline model (second column), and pseudo labels (third column). Boxes indicate examples of high (green), medium (orange), and low (red) performance of non-cropland identification, relating to the detection of (A) large patches of woody vegetation, (B) isolated fields in non-cropland surroundings, (C) isolated trees, (D) clusters of trees, (E) large trees with irregular canopies, and (F) short-term fallow cropland.

# Discussion

We demonstrated how pre-trained field delineation models can be leveraged to generate informative and diverse pseudo labels which facilitate domain adaptation and help fine-tuning models for use in a new target region without the need to train models from scratch. We discuss our findings concerning 1) the high baseline performance of the pre-trained model and the added value of fine-tuning with human labels, 2) opportunities for mobilizing the value of pseudo labels for geographic domain adaptation, 3) and challenges and opportunities for large-area field delineation in smallholder agriculture.

### *Baseline performance & added value of model fine-tuning*

Our experiments are based on a pre-trained FracTAL ResUNet which has been shown to generalize well across regions and outperforms other state-of-the-art deep learning architectures for field delineation (Tetteh et al., 2023). In the heterogeneous setting of the North of Mozambique, the FracTAL ResUNet achieved a good baseline performance in both field delineation and field size estimation (mIoU = 0.634, mRMSE = 0.071). The performance of the pre-trained model was in line, or even exceeding the performance of previous studies in Sub-Saharan Africa, including field delineation in Kenya using Sentinel-2 and PlanetScope data with mIoU scores of up to 0.64 (Kerner et al., 2023), field delineation in Nigeria and Mali using 0.5 m World-View 3 imagery with F1 scores ranging between 0.6 and 0.7 (Persello et al., 2019), or field size estimation in Ghana, where predicted field size was on average 2.91 ha (141%) above reference field size (2.06 ha) (Estes et al., 2022). It is important to note that the absence of a "good practices" protocol outlining key metrics for field delineation hampers direct comparisons of field delineation performance across studies.

While the good baseline performance of the pre-trained model may suggest that satisfying results in increasingly complex landscapes can be achieved without fine-tuning, as also reported by (Wang et al., 2022), this finding should be interpreted as context-specific to our study rather than generalized across all landscapes. The performance of field delineation is strongly dependent on the input imagery and the regional characteristics of the agricultural system, most notably the size of the fields present (Kerner et al., 2023; Nakalembe and Kerner, 2023). Here, we performed inference on images with a spatial resolution of approximately one-third of the training data (0.6 m used here as compared to 1.5 m in the training data), while median field size was around one-quarter of that present in the India data (0.06 ha as compared to 0.24 ha). Wang et al. (2022) highlighted the relevance of spatial resolution for smallholder field delineation by systematically degrading the spatial resolution of Airbus data, finding that performance decreased sharply, particularly affecting the lower end of the field size distribution. Consequently, inference on 1.5 m SPOT data in Mozambique may yield reduced baseline performance for smaller fields as compared to the higher resolution images used here.

Similar to Wang et al. (2022), fine-tuning the India model using data from Mozambique consistently increased model performance across metrics in a comparable range of IoU (our mIoU: +0.05, median IoU in Wang et al.: +0.11) and IoU50 (ours: +0.07, Wang et al. 2022: +0.20), noting that the pre-trained model already yielded satisfactory performance.

*Mobilizing the potential of pseudo labels*

We found that when using pseudo labels, IoU50 increased by up to 86% and the decrease in mRMSE by up to 65% compared to human annotated labels. We explored a variety of methods for filtering pseudo labels using instance-level scores involving semantic confidence, instance confidence, and instance size and tested absolute as well as image-adaptive thresholds for pseudo label selection. The best selection strategy was based on site-level adaptive thresholds for semantic confidence, by which the 1% most confident predictions were selected for each site. Compared to the selection based on absolute thresholds, adaptive selection likely increases the diversity of pseudo labels and thus leads to better generalization (Prabhu et al., 2022). Simultaneously, the adaptive thresholds generated a higher number of pseudo labels as compared to the most conservative absolute threshold. Due to the observed properties in terms of quantities, high spatial agreement, and similar area properties as compared to human labels, a selection based on adaptive thresholds is a suggested option when employing the confidence metrics used here.

The conservative pseudo label selection used here represents a trade-off in terms of quantity, which may be mitigated by running pseudo label selection across larger quantities of unlabeled image chips when these are available. Diversity may further be enhanced by adding sampling strata in site selection to account for agroecological zones (FAO, 2021). In applications where multi-temporal observations or image acquisition dates are accessible and can be included during sampling, the timing of image acquisition may be further accounted for to enhance the temporal generalization of the model. As such, we advise users to use conservative (e.g. 99$^{th}$ percentile) adaptive thresholds for pseudo label selection and control the number and characteristics of the sites included appropriately to obtain optimal model performance.

Self-training using pseudo labels may lead to error propagation in model performance due to the self-referential nature of the approach (Toldo et al., 2020). The introduction of biases in the model is a potential risk in self-learning approaches, which is hard to monitor. Therefore, we recommend a comparison with human-annotated data or a quality screening of pseudo labels for downstream applications. We here observed a tendency of the models to undersegment fields in the region when including pseudo labels of lower confidence, as expressed in substantial declines in Recall. We, therefore, advise against the use of liberal selection criteria relying on absolute thresholds below 0.975, and adaptive thresholds selecting beyond the 1% most confident instances. Moreover, manual screening (i.e. a simple site-level quality assessment of pseudo labels) can be conducted to effectively minimize the risk for erroneous pseudo labels in the training dataset and thus represents a cost-efficient middle ground to produce labels at scale without the need for costly annotation.

Our experiments on complementing human data with pseudo labels for additional sites confirmed that performance increases can be easily achieved by providing diverse, but highly confident pseudo labels for complementary sites. We thus advise the combined use of human-annotated labels and quality-screened pseudo labels wherever possible to obtain optimal performance and reduce the risk of error propagation. Moreover, the use of iterative training may help to overcome the relatively poor discrimination of cropland and non-cropland in the region, which was dominated by overpredictions of cropland in our results.

*Challenges and opportunities for wall-to-fall field delineation*

Two key barriers to wall-to-wall field delineation persist in Sub-Saharan Africa and beyond, namely a lack of labeled data to train deep learning models, and a lack of accessible VHR imagery (Nakalembe and Kerner, 2023). Regarding labels, a compilation of existing labeled datasets in smallholder agriculture from in Rwanda (NASA Harvest et al., 2022), Kenya (Kerner et al., 2023), and Ghana (Estes et al., 2022), may allow for generating a standardized benchmark for testing field delineation approaches across heterogeneous geographies and image datasets. Publicly sharing existing labeled data and model weights, as done by Wang et al. (2022) and in this study, can help to alleviate the existing label bottleneck. Additionally, we propose to consider pseudo labels for producing or complementing training data in label-free or label-scarce settings.

Critically, wall-to-wall image data at appropriate spatial resolution is needed to conduct mapping across large areas. In the case of Northern Mozambique, half of the fields in our reference data were smaller than 0.06 ha, which translates into the need for VHR data at appropriate spatial resolution. Wang et al. (2022) revealed that IoU scores for fields below 0.06 ha size (affecting 11% of the fields in their sample) often had IoU scores below 0.8 when using SPOT data, indicating that ideally, sub-meter spatial resolution data would be needed to capture these small fields. Accessing commercial VHR imagery remains a costly exercise, and local-level or sample-based studies (Lesiv et al., 2019) will remain common trade-offs as long as these cost barriers persist. Alternatively, strengthening public-private initiatives similar to the NICFI data program (Planet Labs Inc., 2023) – enabling public access to monthly PlanetScope mosaics at ~5 m spatial resolution across the global tropics – can aid in overcoming this issue (Estes et al., 2022; Wang et al., 2022). Alternatively, airborne imaging campaigns are available for selected countries and can be leveraged for mapping efforts (e.g. Yin et al. (2023) for Benin), but these remain rare, particularly in Sub-Saharan Africa.

## Conclusion

This study explores the fully automated generation of pseudo labels for smallholder field delineation, which can be used for fine-tuning deep convolutional neural networks across geographies. The workflow is based on publicly available model weights with good capabilities for generating predictions of field boundaries, but is architecture-agnostic and independent of source data. We demonstrated how the "known knowns" can be leveraged to inform on the "known unknowns" and thereby assist in geographic domain adaptation for field delineation.

Our findings highlight the potential of pseudo labels for overcoming domain gaps free of reference data, or their use as a complement to limited human-annotated data, which can further increase model performance. Future research should assess the performance of pseudo labels across multiple smallholder environments, test their production with coarser spatial resolution imagery, the use of alternative instance-level metrics, and investigate ideal balance between human labels and pseudo labels. Further mobilizing the potential of pseudo labels for closing geographic domain gaps is a stepping stone for large-area field delineation in smallholder contexts without the need for costly label collection.

## Acknowledgements

This work was supported by the F.R.S.-FNRS, grant no. T.0154.21 and grant no. 1.B422.24 and the European Research Council (ERC) under the European Union's Horizon 2020 research and innovation program (Grant agreement No 677140 MIDLAND). This research contributes to the Global Land Programme (http://glp.earth). The authors would like to thank the labeling team at Humans in the Loop for organizing and performing label collection.